\documentclass[11pt,journal,cspaper,compsoc,twoside]{IEEEtran}

\usepackage{color,xspace}
\usepackage{amssymb}
\usepackage{graphicx}
\usepackage{hyperref}
\usepackage{amsmath}
\usepackage{amssymb}
\usepackage{booktabs}
\usepackage{calc}
\usepackage{etoolbox}
\usepackage{bigfoot}

\usepackage[export]{adjustbox}[2011/08/13]

\newtoggle{tech_report}
\newtoggle{review_copy}
\newtoggle{tight}

\toggletrue{tech_report}

\iftoggle{review_copy}{%
\usepackage{ruler}
}

\usepackage{url}
\urldef{\mailsa}\path|{ken, karen, az}@robots.ox.ac.uk|  

\newcommand{\oneDot}{.\xspace}

\def\eg{\emph{e.g}\oneDot}

\def\ie{\emph{i.e}\oneDot}
\def\vs{\emph{vs}\oneDot}

\newcommand{\bw}{\mathbf{w}}

\newcommand{\enc}{\phi}

\newcommand{\rk}[1]{{\bf\color{blue} #1}}
\newcommand{\image}{I}

\newcommand{\sgn}{\operatorname{sgn}}
\DeclareMathOperator{\Rbb}{\mathbb{R}}

\begin{document}

\nottoggle{tech_report}{%
\input{splncs_preamble.tex}
}{%
\renewenvironment{figure}{\begin{figure*}}{%
    \end{figure*}\ignorespacesafterend
}
\renewenvironment{table}{\begin{table*}}{%
    \end{table*}\ignorespacesafterend
}
\renewenvironment{paragraph}[1]{\par\emph{#1}}{}

\newcommand{\widefigure}{1.0}
\newcommand{\narrowfigure}{0.8}
\newcommand{\pagefigure}{0.7}


\title{Efficient On-the-fly Category Retrieval \\using ConvNets and GPUs}

\author{Ken~Chatfield, Karen~Simonyan, and~Andrew~Zisserman\\
        Visual Geometry Group, Department of Engineering Science, University of Oxford \\        
        {\small \{ken,karen,az\}@robots.ox.ac.uk}
}
        

\IEEEcompsoctitleabstractindextext{%
\begin{abstract}
We investigate the gains in precision and speed, that can be obtained by using
Convolutional Networks (ConvNets) for on-the-fly retrieval -- where classifiers
are learnt at run time for a textual query from downloaded images, and used to rank
large image or video datasets.

We make three contributions: (i) we present an evaluation of state-of-the-art
image representations for object category retrieval over standard benchmark
datasets containing 1M+ images; (ii) we show that ConvNets can be used to
obtain features which are incredibly performant, and yet much lower dimensional
than previous state-of-the-art image representations, and that their dimensionality
can be reduced further without loss in performance by compression using product
quantization or binarization. Consequently, features with the state-of-the-art
performance on large-scale datasets of millions of images can fit in the
memory of even a commodity GPU card; (iii) we show that an SVM classifier can be
learnt within a ConvNet framework on a GPU \emph{in parallel} with downloading the
new training images, allowing for a continuous refinement of the model
as more images become available, and simultaneous training and ranking.
The outcome is an on-the-fly system that significantly
outperforms its predecessors in terms of: precision of retrieval,
memory requirements, and speed, facilitating accurate on-the-fly learning and ranking
in under a second on a single GPU.
\end{abstract}}

\maketitle

\IEEEdisplaynotcompsoctitleabstractindextext
}%


\section{Introduction}\label{sec:intro}

\nottoggle{tech_report}{%
On-the-fly
}{%
\IEEEPARstart{O}{n-the-fly}
}%
learning offers a way to overcome the `closed world'
problem in computer vision, where object category recognition systems
are restricted to only those pre-defined classes 
that occur in the carefully curated
datasets available for training 
-- for example ImageNet~\cite{Deng09} for object
categories or UCF-101~\cite{Soomro12} for human actions in videos.
What is more, it offers the tantalising prospect of developing
large-scale general purpose object category retrieval systems which
can operate over millions of images in a few seconds, as is possible
in the specific instance retrieval
systems~\cite{Sivic03,Nister06,Philbin07,Jegou08,Jegou10} which have
reached the point of commercialisation in products such as Google
Goggles, Kooaba and Amazon's SnapTell.

Current on-the-fly systems typically proceed in three
stages~\cite{Chatfield12,Fernando13,Liu09a,Parkhi12b}: first, training data for
the user query are compiled, commonly by bootstrapping the process via
text-to-image search using \eg Google Image Search as a source of training images;
second, a classifier or ranker is learnt for that category; third, all
images/videos in a dataset are ranked in order to retrieve those containing the
category. The aim is for these stages to happen on-line in a matter of seconds,
rather than hours.

Previous methods for on-the-fly learning have been limited by the
retrieval-performance/memory/speed trade off. In particular, very high-dimensional
feature vectors were required for state-of-the-art classification
performance~\cite{Chatfield11,Perronnin10a,Perronnin12}, but this
incurred both a severe memory penalty (as features for the dataset need
to be in memory for fast retrieval) and also a severe speed penalty
(as computing a scalar product for high-dimensional features is
costly) both in training and ranking. Despite the excellent
progress in compression methods for nearest neighbour search by using
product quantization~\cite{Jegou11} or binary encoding~\cite{Torralba08,Raginsky09}, 
compromises still had to be made. 

\iftoggle{tech_report}{%
\begin{figure}[t]
    \makebox[\textwidth][c]{\includegraphics[width=\narrowfigure\textwidth]{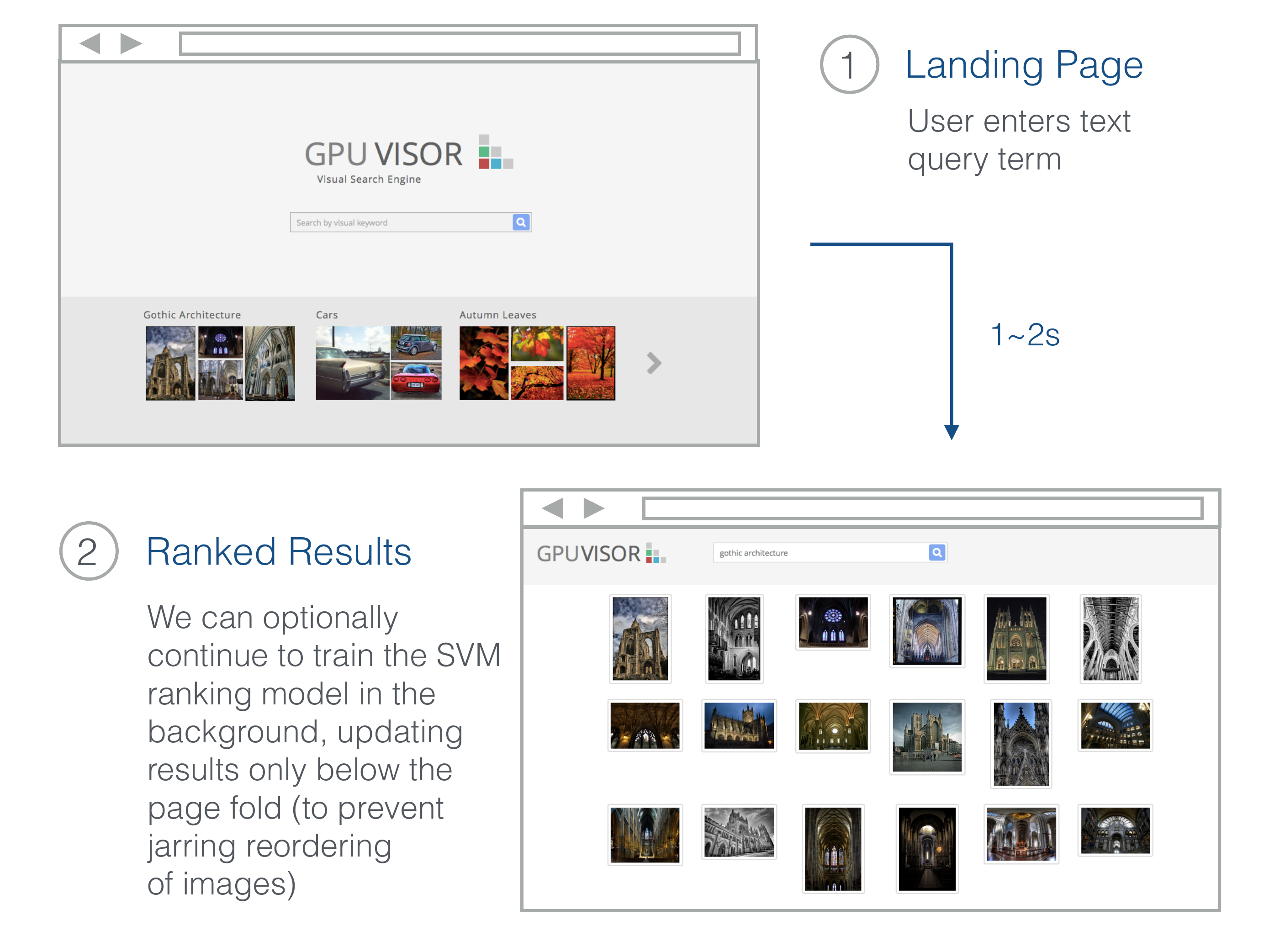}}
    \caption{\textbf{Live on-the-fly system web frontend}. From entering a novel text query to viewing
    the results page, the entire process takes $\sim$1--2 seconds.}
    \label{fig:gpu-visor}
    \iftoggle{tight}{\vspace{0.2em}}{}
\end{figure}
}{%
\begin{figure}[t]
    \makebox[\textwidth][c]{\includegraphics[width=\narrowfigure\textwidth]{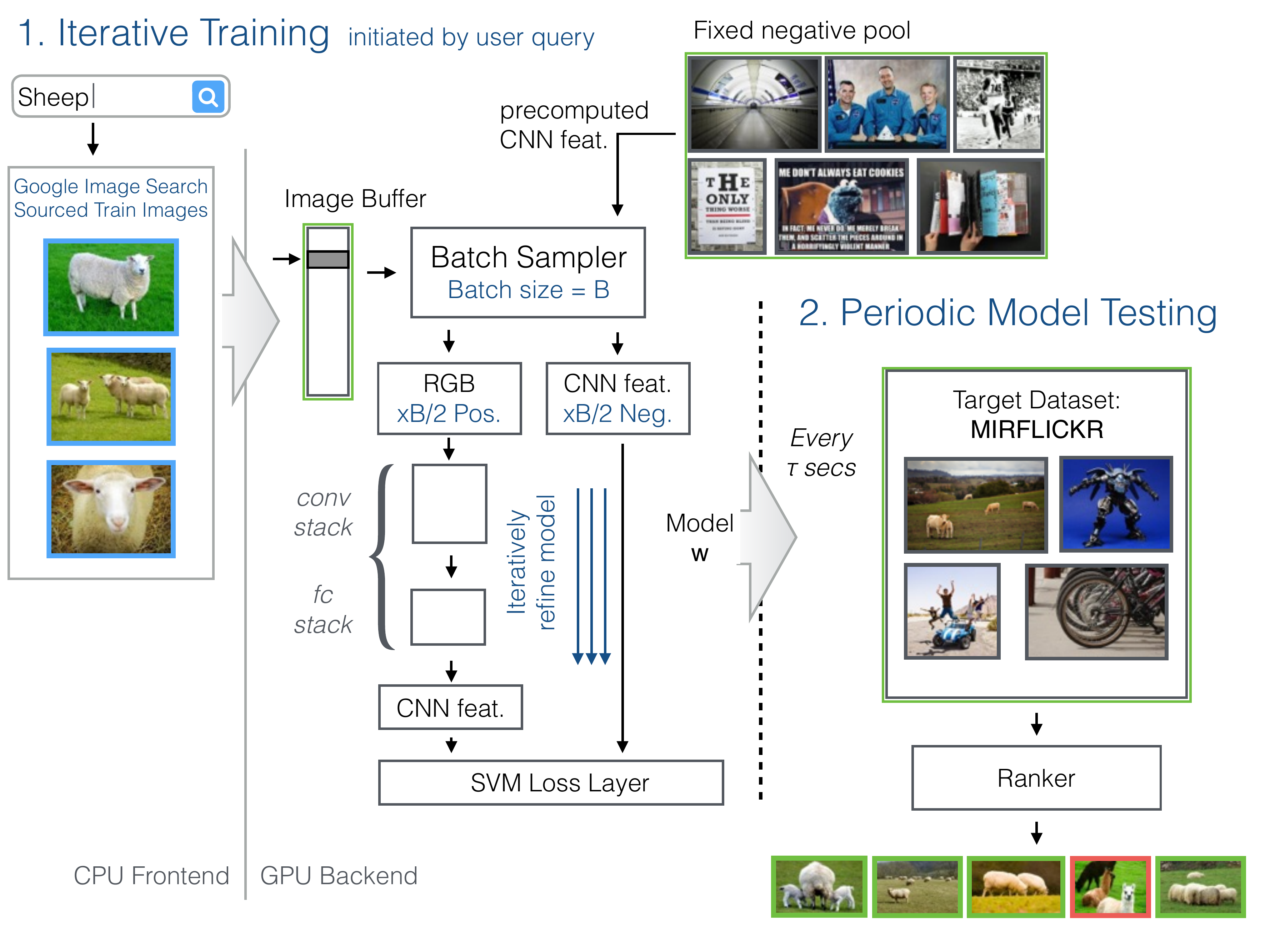}}
    \caption{\textbf{Architecture of our on-the-fly object category retrieval system.} The entire framework
    aside from the image downloader is resident on the GPU, with data stored in GPU memory outlined in
    green. Its operation is split into: (i) iterative training, as initiated by a user text
    query and (ii) periodic model testing to obtain a ranking over the target dataset (refer to
    text for further details).}
    \label{fig:architecture}
\end{figure}
}

In this paper we show that in the context of on-the-fly category retrieval,
Convolutional Networks (ConvNets)~\cite{Lecun98} with GPU training~\cite{Krizhevsky12} can
significantly improve on all three of: retrieval precision, memory requirements, and ranking speed.
The whole pipeline, from computing the training image features and learning the model to
scoring and ranking the dataset images is implemented on the GPU and runs in a highly-parallel, online
manner.
We thus demonstrate a system that is able to go from a cold-query to results in a matter of
second(s) on a dataset of million(s) of images%
\iftoggle{tech_report}{%
~(some screenshots of our live on-the-fly retrieval
system and web frontend are shown in Figure~\ref{fig:gpu-visor}).
}{%
. The architecture of our proposed system, from input of text query
to display of ranked results, is summarized in Figure~\ref{fig:architecture} (refer to Section~\ref{sec:arch}
for details).
}

In terms of retrieval performance, we build on the recent research that
shows that deep ConvNet features significantly outperform shallow features,
such as Fisher Vectors~\cite{Perronnin10a,Chatfield11}, on the image classification 
task~\cite{Krizhevsky12,Zeiler13,Chatfield14}.
However, our contributions go further than simply using ConvNet features in an
on-the-fly architecture: we take full advantage of the GPU computation for all
retrieval stages%
\iftoggle{tech_report}{%
, in parallel with downloading the new training images on the CPU.
}{%
.
}
This novel GPU-based architecture allows a time budget to be set, so that an SVM, trained on the available images 
within the time limit, can be used to (re-)rank the dataset images at any stage of the process (for instance, every $0.5$s).
This architecture is in strong contrast to the standard on-the-fly architectures~\cite{Chatfield12}, where SVM
training only begins once all training images have been downloaded and processed, and ranking follows after that.


We start by conducting a comprehensive evaluation
of the performance of ConvNet-based image features for category-based image
retrieval. Given the lack of evaluation data suitable for the assessment
of large-scale retrieval performance, we compose our own by taking a
standard medium-scale object category recognition benchmark
(PASCAL VOC 2007~\cite{Everingham10}) and then optionally adding a large
number of distractor images to take the dataset size to 1M+ images.
We evaluate over these two datasets under variation in training data --
either using VOC training images (\ie~a curated dataset) or using images
from Google Image search (\ie~the type of images, possibly with label
noise, that will be available for the real-world on-the-fly system).
Full details are given in Section~\ref{sec:evdesc}.

With our goal being ranking of millions of images on a conventional GPU-equipped PC, we
then investigate, in Section~\ref{sec:retrieval},
how retrieval performance is affected
by using low-dimensional features (still originating from a
ConvNet) over these scenarios. Low-dimensional features (e.g.\ hundreds of components
rather than thousands) have two advantages: they use less memory, and
scalar products are faster, both in training and ranking. We cover a
spectrum of methods for achieving a low-dimensional descriptor, namely:
(i)~reducing the dimensionality of the last \mbox{ConvNet} layer;
(ii)~product quantization of the ConvNet features and 
(iii)~binarization of the ConvNet features.  
It is shown that a combination of a low-dimensional final ConvNet
feature layer with product quantization produces features that are both 
highly-compact and incredibly performant.

Finally, based on these investigations, we propose a GPU architecture
for on-the-fly object category retrieval in Section~\ref{sec:arch},
highly scalable, capable of adapting to varying query complexity
and all running on a single commodity GPU.
\iftoggle{tech_report}{%
}{%
An extended version of this paper is available on arXiv\footnote{\url{http://arxiv.org/abs/1407.4764/}}.
}

%
%

\section{Evaluating Large-scale Object Category Retrieval}\label{sec:evdesc}

This section describes the evaluation protocol used to assess the
performance of the image representations $\enc(\image)$ described in
Section~\ref{sec:retrieval} and of the on-the-fly training architecture
introduced in Section~\ref{sec:arch}. We begin by describing the
datasets used for evaluation, and then describe the three different
scenarios in which these datasets are used, with each subsequent
scenario moving closer to modelling the conditions experienced by a
real-world large-scale object category retrieval system.
%

One difficulty of evaluating a large-scale object category retrieval
system is the lack of large-scale datasets with sufficient annotation
to assess retrieval performance fully, in particular to measure
recall. The PASCAL VOC
dataset~\cite{Everingham10} provides full annotation for a set of
twenty common object classes, facilitating evaluation using common
ranking performance measures such as mean average precision (mAP), but
is much too small ($\sim$10k images) to evaluate the performance of a
real-world system.
Conversely, the ILSVRC dataset~\cite{Deng09}, while being much
larger ($\sim$1M+ images), does not have complete annotation of \emph{all} object categories in each image.
Therefore, ranking performance (\eg~recall or mAP) cannot be measured without further
annotation, and only object category \emph{classification}
metrics (such as top-N classification error per image), which do not accurately reflect
the performance of an object category \emph{retrieval} scenario, can be used.
Additionally, in this work we use the ImageNet ILSVRC-2012 dataset to pre-train
the ConvNet, so can not also use that for assessing
\iftoggle{tech_report}{%
retrieval performance.
}{%
performance.
}%

As a result, for evaluation in this paper, we use a custom combination of datasets,
carefully tailored to be representative of the data that could be expected in a typical
collection of web-based consumer photographs:
\iftoggle{tight}{\vspace{-0.7em}}{}
\paragraph{PASCAL VOC 2007}~\cite{Everingham10} is used as our base dataset,
with assessment over seventeen of its twenty classes (`people',
`cats' and `birds' are excluded for reasons explained below). 
We use the provided train, validation and test splits.
\iftoggle{tight}{\vspace{-0.7em}}{}
\paragraph{MIRFLICKR-1M}~\cite{Huiskes08,Huiskes10} 
is used to augment the data from the PASCAL VOC~2007 test set in our
later experiments, and comprises 1M unannotated images (aside from
quite noisy image tags). The dataset represents a snapshot of images
taken by popularity from the image sharing site Flickr, and thus is more
representative of typical web-based consumer photography than \mbox{ImageNet},
which although also sourced from Flickr was collected through queries for
often very specific terms from WordNet. In addition, MIRFLICKR-1M has been
confirmed to contain many images of the twenty PASCAL VOC classes.

\subsection{Evaluation Protocol}\label{sec:evproto}

\iftoggle{tech_report}{%
A linear SVM is trained for all classes, and used to rank all images
in the target dataset. We are interested in evaluating the performance within
an object category retrieval setting, and so measuring the `goodness'
of the first few pages of retrieved results is critical. We therefore
evaluate using precision @ $K$, where $K=100$, on the basis that the
larger the proportion of true positives for a given object category at
the top of a ranked list the better the perceived performance.
}{%
A linear SVM is trained for all classes, and used to rank all images
in the target dataset. For the object category retrieval setting the `goodness' of the first few pages of retrieved
results is critical, as the
larger the proportion of true positives for a given object category at
the top of a ranked list, the better the perceived performance.
We therefore
evaluate using precision @ $K$, where $K=100$.
}

Adopting such an evaluation protocol also has the advantage that we
are able to use the 1M images from the MIRFLICKR-1M dataset despite
the fact that full annotations are not provided. Since we only need
to consider the top $K$ of the ranked list for each class during
evaluation, we take can take a `lazy' approach to annotating the
MIRFLICKR-1M dataset, annotating class instances
only as far down the ranked list as necessary to generate a complete
annotation for the top-$K$ results (for more details of this
procedure, refer to scenario 2 below). This
avoids having to generate a full set of annotation for all 1M
images.

\subsection{Experimental Scenarios}\label{sec:scenarios}

\iftoggle{tech_report}{%
\begin{figure}[t]
    \makebox[\textwidth][c]{\includegraphics[width=\narrowfigure\textwidth]{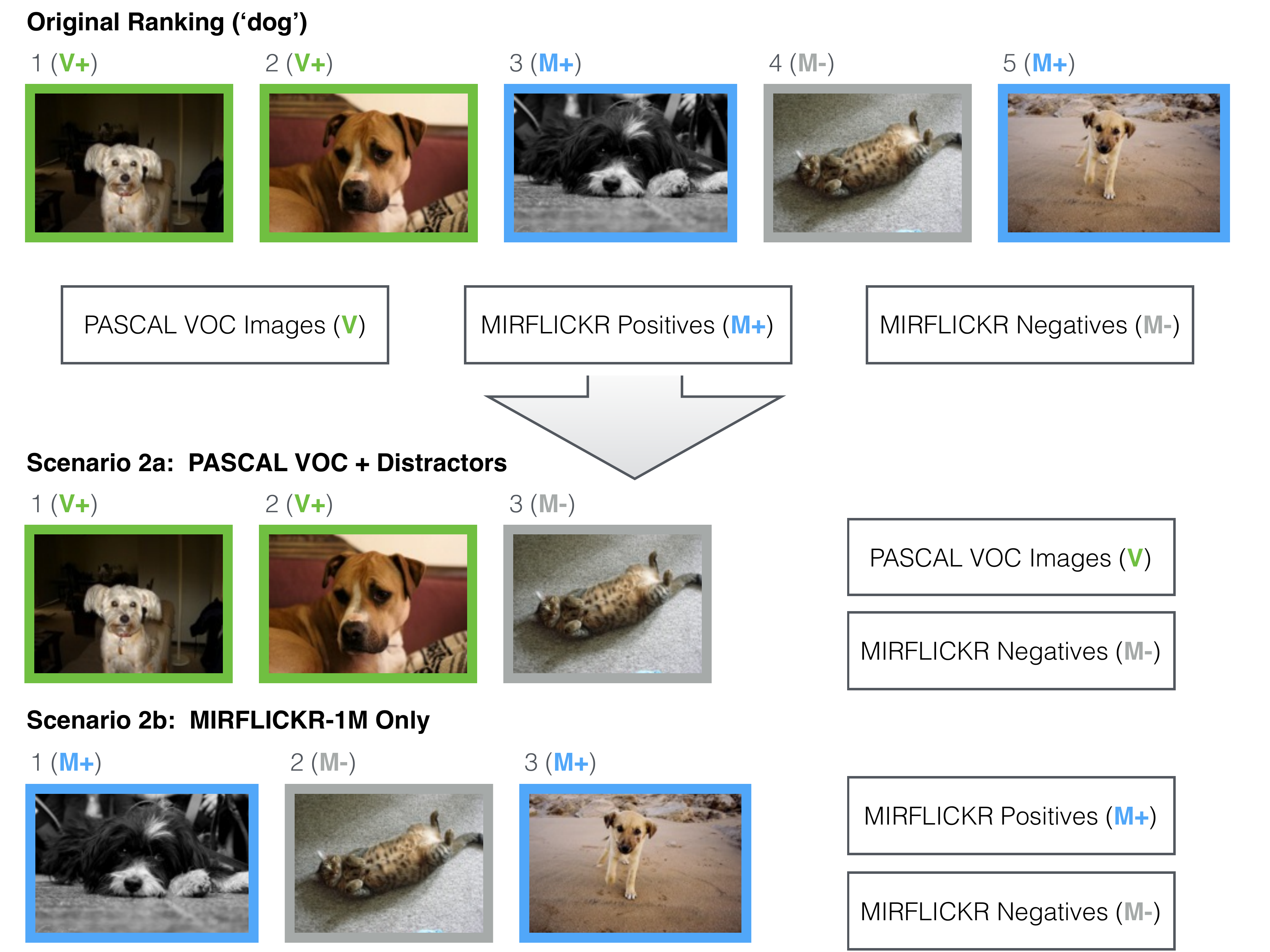}}
    \caption{\textbf{Data subsets used for evaluation.}
    Using the example of the object category `dog', the ranked lists used for evaluation in scenarios 2a+2b
    are compiled by combining the PASCAL VOC data with lazily annotated data from
    the MIRFLICKR-1M dataset.}
    \label{fig:evaluation-rankings}
    \iftoggle{tight}{\vspace{0.2em}}{}
\end{figure}
}{}

\subsubsection{Scenario 1: PASCAL VOC.} 
We train models for seventeen of the twenty VOC object classes
(excluding `people', `cats' and `birds') using both the training and
validation sets. Following this, a ranked list for each class is
generated using images from the test set and precision @ $K$ evaluated.
\iftoggle{tight}{\vspace{-0.7em}}{}
\subsubsection{Scenario 2: Large-scale Retrieval.} 
Training is undertaken in the same manner as scenario 1, but during
testing images are added from the \mbox{MIRFLICKR-1M} dataset.  There are two
sub-scenarios%
\iftoggle{tech_report}{%
~(using different subsets of the test data, summarised in
Figure~\ref{fig:evaluation-rankings}).
}{%
:
}
\iftoggle{tight}{\vspace{-0.7em}}{}
\paragraph{Scenario 2a --} 
we test using images from the PASCAL VOC test set (as in scenario
1) with the addition of the entirety of the MIRFLICKR-1M dataset.  For
each class,
\iftoggle{tech_report}{%
we remove all positive class occurrences in the ranked
list which are retrieved from the MIRFLICKR-1M dataset
using the lazy annotation described in Section~\ref{sec:gt},
}{%
we remove all (lazily annotated) positive class occurrences in the ranked
list which are retrieved from MIRFLICKR-1M,
}%
as the purpose of
this scenario is to test how our features perform when attempting to
retrieve a small, known number of class occurrences from a very large
number of non-class `distractor' images.\footnote{The prevalence of the PASCAL VOC classes `people', `cats' and `birds' in the MIRFLICKR-1M data explains why we exclude them, as restricting the annotation of these classes to reasonable levels proved to be impossible.}
\iftoggle{tight}{\vspace{-0.7em}}{}
\paragraph{Scenario 2b --} 
this time we exclude all images from the PASCAL VOC dataset, and
instead evaluate precision @ $K$ solely over the MIRFLICKR-1M
\iftoggle{tech_report}{%
dataset, lazily annotating the retrieved ranked lists in each case as
before.
}{%
dataset.
}%
The purpose of this scenario is to test how our features
perform over a real-world dataset with unknown statistics. In
practice, it is an easier scenario than scenario 2a, since the
MIRFLICKR-1M dataset contains many instances of all of the PASCAL VOC
classes.
\iftoggle{tight}{\vspace{-0.7em}}{}
\subsubsection{Scenario 3: Google Training.} 
Testing is the same as in scenario 2b, but instead of using PASCAL data
for training, a query is issued to Google Image search for each of the
PASCAL VOC classes, and the top $N\sim250$ images are used in each
case as training data.  This scenario assesses the tolerance to
training on images that differ from the VOC and MIRFLICKR-1M test
images: the Google images may be noisy and typically contain the object
in the centre.
It also mirrors most
closely a real-world on-the-fly object category retrieval setting, as
the queries in practice do not need to be limited to the PASCAL VOC
classes.  There are again two sub-scenarios, with different data used
for the negative training samples in each case:
\paragraph{Scenario 3a --} 
the images downloaded from Google Image Search for all other classes,
except for the current class, are used as negative training data (this
mirrors the PASCAL VOC setup).
\iftoggle{tight}{\vspace{-0.7em}}{}
\paragraph{Scenario 3b --} 
a fixed pool of $\sim16,000$ negative training images is used. These
training images are sourced from the web by issuing queries for a
set of fixed `negative' query terms\footnote{miscellanea, random selection,
photo random selection, random objects, random things, nothing in particular,
photos of stuff, random photos, random stuff, things}
to both Google and Bing image search,
and attempting to download the first $\sim 1,000$ results in each case.
This same pool of negative training data is also used in Section~\ref{sec:arch}.



\iftoggle{tech_report}{%
\subsection{Dataset Ground Truth Preparation}\label{sec:gt}

As described in Section~\ref{sec:scenarios}, we use a combination of the PASCAL VOC 2007 dataset
and MIRFLICKR1M dataset for evaluation. MIRFLICKR1M does not come with any annotation, apart from noisy flickr image tags, and so we add our own annotations for the twenty PASCAL VOC classes.

Despite the dataset containing 1M images, we can get away with annotating far less than this
number given our chosen evaluation metric, precision @ $K$ with $K=100$, which only requires
the ground truth for the first $K$ items in the ranked list of each target class to compute.
We therefore adopt a `lazy' approach to annotation using our result ranked lists as a
starting point.

The evaluation set (and thus the meaning of the `first $K$ images') is different for
each scenario, as shown in Figure 1 of the paper. Therefore, given any raw ranked list
for class $C$ (which is a combination of results from both the PASCAL VOC and
MIRFLICKR1M datasets) it suffices to annotate images which fall within the following
ranges:

\begin{itemize}
\item \textbf{For Scenario 2a --} the top $K$ images from: the PASCAL VOC dataset combined with
all images from MIRFLICKR1M annotated as negative for the target class $C$
(these are `distractors')\\\ldots%
\emph{excluding} annotated positives for class $C$ from MIRFLICKR1M.
\item \textbf{For Scenario 2b/3 --} the top $K$ images from the MIRFLICKR1M dataset%
\\\ldots\emph{excluding} all PASCAL VOC images.
\end{itemize}

The annotations we make for any particular method/scenario should be stored so that
images do not need to be annotated more than once for different methods.
We developed a web-based
annotation tool to facilitate this process
which allows positive
annotations for a class to be shared across both methods and scenarios.

In total, 46,770 images from the MIRFLICKR1M dataset were annotated, with an average
of $\sim2,000$ annotations per class. These annotations will be made publicly
available.

}{}

\section{Retrieval Performance over Image Representations}\label{sec:retrieval}

In this section, we perform an evaluation of recent state-of-the-art
image representations for the object category retrieval scenarios described
in Section~\ref{sec:scenarios}.

ConvNet-based features, which form the basis of our on-the-fly system described in
Section~\ref{sec:arch}, have been shown to perform excellently on standard
image classification benchmarks such as PASCAL VOC and ImageNet
ILSVRC~\cite{Zeiler13,Donahue13} \cite{Chatfield14,Razavian14}. We therefore focus
our evaluation on these features, employing 2048-dimensional `CNN M 2048' image features of~\cite{Chatfield14}
as the baseline. We compare them to a more traditional shallow feature encoding in the form of the Improved Fisher Vector
(IFV)~\cite{Perronnin10a}. Implementation details for ConvNets and IFV are given in Section~\ref{sec:implement}.
We explore the effects of reducing the dimensionality of our features on their retrieval performance using the following methods:
\iftoggle{tight}{\vspace{-0.7em}}{}
\paragraph{Lower-dimensional ConvNet output layer --}
One way of reducing the dimensionality of ConvNet features consists in retraining the network so that the last fully-connected (feature) layer has a
lower dimensionality. Following~\cite{Chatfield14}, we consider the `CNN M 128' network configuration with a 128-dimensional feature layer.
Using such network in place of the baseline `CNN M 2048' can be seen as discriminative dimensionality reduction by a factor of 16.
\iftoggle{tight}{\vspace{-0.7em}}{}
\paragraph{Product quantization (PQ)}
has been widely used as a compression method for image features~\cite{Jegou11,Sanchez11}, 
and works by splitting the original feature into $Q$-dimensional
sub-blocks, each of which is encoded using a separate vocabulary of cluster centres pre-learned from
a training set.
Here we explore compression using $Q=4,8$-dimensional sub-blocks.
\iftoggle{tight}{\vspace{-0.7em}}{}
\paragraph{Binarization}
is performed using the tight frame expansion method of~\cite{Jegou12b}, which has been recently successfully applied to local patch and face descriptors~\cite{Simonyan14,Parkhi14}.
The binarization of zero-centred descriptors $\enc \in \Rbb^m$ to binary codes $\beta \in \{0,1\}^n, \; n>m$ is performed
\iftoggle{tech_report}{%
as follows:
\begin{equation}
\label{eq:binarize}
\beta = \sgn(U\enc)
\end{equation}
}{%
as $\beta = \sgn(U\enc)$
}%
where is the sign function: $\sgn(a)=1$ iff $a>0$ and $0$ otherwise, and the Parseval tight frame 
$U$ is computed by keeping the first $m$ columns of an orthogonal matrix, obtained from a QR-decomposition of a random $n \times n$ matrix.


\vspace{1em}
\subsection{Results and Analysis}\label{sec:eval}

\begin{figure}[t]
    \iftoggle{tech_report}{%
    \makebox[\textwidth][c]{\includegraphics[width=\narrowfigure\textwidth]{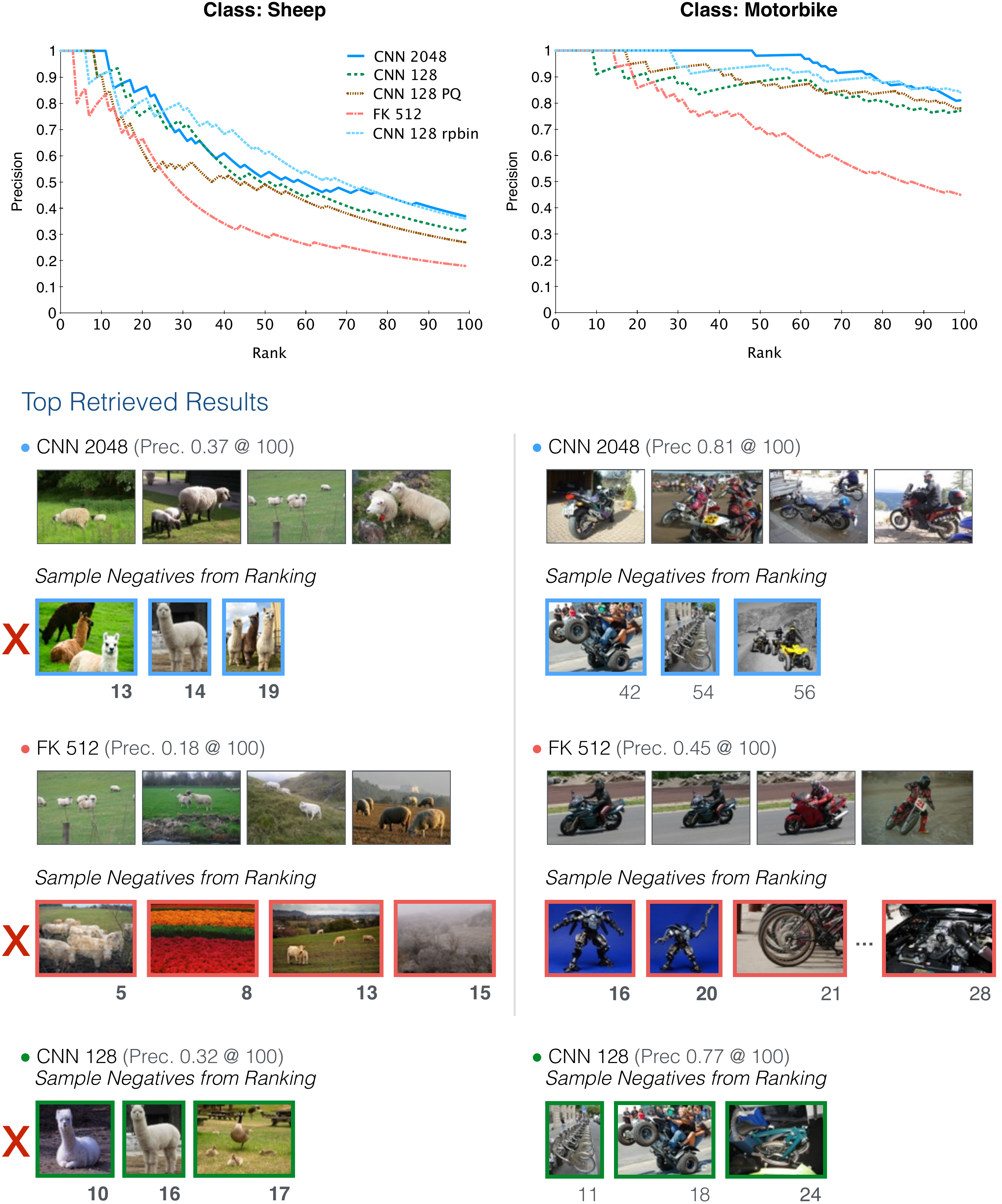}}
    \caption{Sample precision-rank curves and retrieved results for two queries `sheep' and `motorbike' over the combined VOC+MIRFLICKR data (Scenario 2a).}
    }{%
    \makebox[\textwidth][c]{\includegraphics[width=\narrowfigure\textwidth]{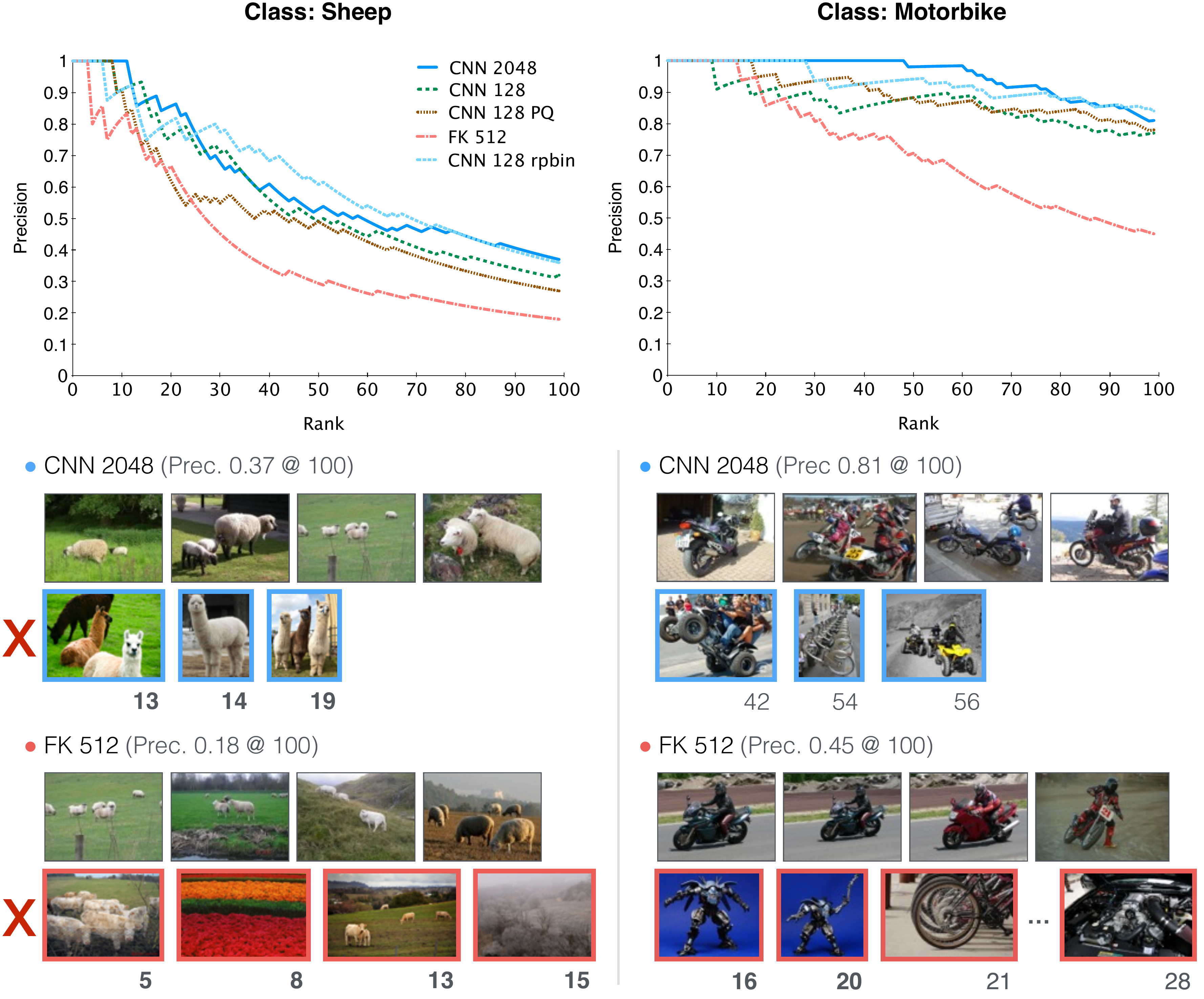}}
    \caption{Sample precision-rank curves and retrieved results for two queries over the combined VOC+MIRFLICKR data (Scenario 2a). In the bottom half of the figure, the top row in each case shows the first few results returned for each method and the second shows the top retrieved false positives with their rank.}
    }%
    \label{fig:pr-rankings}
\end{figure}

\iftoggle{tech_report}{%
The results for all three experimental scenarios are presented in Table~\ref{tab:res_tab}.
We discuss the results for each scenario below.
}{%
}%

\subsubsection{Scenario 1  (VOC Train/VOC Test).}
The PASCAL VOC dataset does not pose any major challenges for any of our features, which
is not surprising given the close to decade of research on representations which perform
well on this dataset.
Even for the most challenging classes (\eg `potted plant') IFV produces
fairly good results, with the top 12 images being true positives (Prec @ 100 = 0.58), and the
top 92 images being true positives in the case of our 2048-dimensional ConvNet features (Prec @ 100 = 0.83).
\iftoggle{tight}{\vspace{-0.7em}}{}
\subsubsection{Scenario 2a (VOC Train/VOC+distractors Test).}
Adding 1M distractor images from the MIRFLICKR-1M dataset has a significant impact on
the results, with the task now being to retrieve true positives that constitute less than $\sim 0.02\%$
of the dataset. This is a more challenging scenario, and under this setting the
superior performance of the ConvNet-based features, when compared to the state-of-the-art shallow
representation (IFV), is much clearer to see.
Some sample precision-rank curves for two queries, one particularly challenging (`sheep') and
another less so (`motorbike') are shown in Figure~\ref{fig:pr-rankings}. We can make the following
observations:

\iftoggle{tight}{\vspace{-0.7em}}{}
\paragraph{IFV Performance --}
It can be seen that IFV~(\ref{exp:fk} in Table~\ref{tab:res_tab})
performs the worst of all methods, despite being much higher dimensional ($\sim 1000\times$)
and taking much longer to compute ($\sim 200\times$) compared to our CNN-128 method~(\ref{exp:cnn_128}). Nonetheless,
even for challenging classes such as `sheep' IFV manages to pull out a few true positives at the
top of the ranked list. However, the relative performance drop with rank is much sharper than with the
ConvNet-based methods.
\iftoggle{tight}{\vspace{-0.7em}}{}
\paragraph{Bursty Images --} comparing the top-ranked
negatives of the FK-512 method~(\ref{exp:fk}) for `sheep' to those of the CNN-2048 method~(\ref{exp:cnn_2k}), it can be seen
that IFV appears to mistakenly rank highly `bursty' images comprising repeating patterns or
textures. This phenomenon is particularly evident for natural, outdoor scenes which explains
why the performance drop of IFV is particularly severe in the `sheep', `cow' and `horses' classes,
as it appears that the ConvNet-based features are much more robust to such textured images%
\iftoggle{tech_report}{%
, although 
the use of heavy PQ compression (\eg the CNN-128-PQ-8 method~\ref{exp:cnn_128_pq8}) starts to show some deterioration as 
a consequence of the retrieval of a smaller number of similarly `bursty' images.
}{%
.
}

\definecolor{grey}{rgb}{0.72,0.7,0.74}
\def\mybar#1{
  {\color{grey}\rule{#1cm}{8pt}}}
  
\renewcommand{\arraystretch}{1.2}

\newcounter{expno}
\renewcommand{\theexpno}{[\alph{expno}]}
\setcounter{expno}{0}

\begin{table}[t]\small
\iftoggle{tech_report}{%
\setlength{\tabcolsep}{2pt}
}{%
\setlength{\tabcolsep}{1pt}
}%
\small
\iftoggle{tight}{\vspace{-0.7em}}{}
\begin{center}
\begin{adjustbox}{center}
\begin{tabular}{@{\extracolsep{10pt}}llllllll@{}}\toprule
       & & \textbf{VOC Only} & \multicolumn{2}{l}{\textbf{Large-scale Retr.}} & \multicolumn{2}{l}{\textbf{Google Training}}\\
Scenario & & \textbf{[1]}      & \textbf{[2a]} & \textbf{[2b]} & \textbf{[3a]} & \textbf{[3b]} \\ \midrule
\refstepcounter{expno}\label{exp:fk}{\small (\alph{expno}) FK} & 512 & 82.3 & 29.3 & 80.5\\ \midrule
\refstepcounter{expno}\label{exp:cnn_2k}{\small (\alph{expno}) CNN} & 2K & 92.1 & \textbf{55.4} & 95.4 & 88.5 & 90.9\\
\refstepcounter{expno}\label{exp:cnn_2k_pq}{\small (\alph{expno}) CNN} & 2K PQ & 90.7 & 55.1 & 96.4 & 88.2 & 91.9\\ \midrule
\refstepcounter{expno}\label{exp:cnn_128}{\small (\alph{expno}) CNN} & 128 & 92.1 & 51.0 & 95.1 & 88.1 & 92.3\\
\refstepcounter{expno}\label{exp:cnn_128_na}{\small (\alph{expno}) CNN} & 128 noaug & 88.8 & 45.4 & 93.1 & 87.1 & 91.1\\
\refstepcounter{expno}\label{exp:cnn_128_rpbin2k}{\small (\alph{expno}) CNN} & 128 BIN 2K & 91.5 & \textbf{52.3} & 94.0 & 89.6\\
\refstepcounter{expno}\label{exp:cnn_128_rpbin1k}{\small (\alph{expno}) CNN} & 128 BIN 1K & 90.0 & 50.1 & 94.0 & 89.5\\
\refstepcounter{expno}\label{exp:cnn_128_pq}{\small (\alph{expno}) CNN} & 128 PQ & 90.1 & 50.5 & 94.6 & 88.2 & 92.1\\
\refstepcounter{expno}\label{exp:cnn_128_pq8}{\small (\alph{expno}) CNN} & 128 PQ-8 & 88.8 & 47.4 & 93.1 & 87.7 & 91.1\\ \bottomrule
\end{tabular}
\end{adjustbox}
\end{center}
\caption{Retrieval results (Mean Prec @ 100) for the evaluation scenarios described in section~\ref{sec:scenarios}.}
\label{tab:res_tab}
\end{table}

\setcounter{expno}{0}

\iftoggle{tight}{\vspace{-0.7em}}{}
\paragraph{Diversity --}
The diversity of the retrieved results is also much greater for ConvNet-based representations than for
IFV, indicating that the classifier is able to make better generalisations using these features.
For example, as seen in Figure~\ref{fig:pr-rankings}, whereas the top four retrieved results for the
query `motorbike' for the FK-512 method~(\ref{exp:fk}) all show a rider in a similar pose, on a racing bike on
a race track, the top four retrieved results for the CNN-2048 method~(\ref{exp:cnn_2k}) depict a variety of different
motorcycles (road, racing, off-road) from several different angles.

For the most part, compression of the ConvNet features does not appear to reduce their diversity
appreciably, with the top-ranked results for all ConvNet methods, whether compressed or not,
appearing to exhibit a similar diversity of results.

\iftoggle{tight}{\vspace{-0.7em}}{}
\paragraph{Compression --}
As mentioned above, the drop in performance in moving from
ConvNet-based features to IFV is much greater than that incurred by
any of the compression methods, and this seems to be strongly
connected with the robustness of the ConvNet-based features, whether
compressed or not, to the kind of `bursty' textured images which IFV
is susceptible to. This is remarkable given that comparing the
size of the largest uncompressed ConvNet representation CNN-2048~(\ref{exp:cnn_2k})
to the smallest PQ-compressed one, CNN-128-PQ-8~(\ref{exp:cnn_128_pq8}), there is
a $\sim 512 \times$ size difference. In the case of the CNN-128-BIN-2K
method~(\ref{exp:cnn_128_rpbin2k}), the mPrec @ 100 actually increases marginally when
compared to the non-compressed codes~(\ref{exp:cnn_128}) which, when visually inspecting
the rankings, again can be explained by the additional robustness brought by compression.

\begin{table}[t]\small
\iftoggle{tech_report}{%
\setlength{\tabcolsep}{2pt}
}{%
\setlength{\tabcolsep}{1pt}
}%
\small
\begin{center}
\begin{adjustbox}{center}
\begin{tabular}{@{\extracolsep{10pt}}llllllll\iftoggle{tech_report}{l}{}@{}}\toprule
 & Dim & \iftoggle{tech_report}{\multicolumn{3}{c}{Compression}}{\multicolumn{2}{c}{Compression}} & New Dim & Storage & Comp. Time\\
  &  & \iftoggle{tech_report}{\multicolumn{3}{l}{}}{\multicolumn{2}{l}{}} & (bytes) & / 1M ims. & / im (s)\\ \midrule
\refstepcounter{expno}{\small (\alph{expno})} & 83,968 & -- & & \iftoggle{tech_report}{&}{} & 312.8 GB & 10.32\\ \midrule
\refstepcounter{expno}{\small (\alph{expno})} & 2048 & -- & & \iftoggle{tech_report}{&}{} & 7.63 GB & 0.35 (0.061)\\
\refstepcounter{expno}{\small (\alph{expno})} & 2048 & PQ & 4 dims/sq \iftoggle{tech_report}{&}{} (16$\times$) & 512 & 488 MB & + 0.061\\ \midrule
\refstepcounter{expno}{\small (\alph{expno})} & 128 & -- & & \iftoggle{tech_report}{&}{} & 488 MB & 0.34 (0.061)\\
\refstepcounter{expno}{\small (\alph{expno})} & 128 & noaug & & \iftoggle{tech_report}{&}{} & 488 MB & 0.083 (\textbf{0.024})\\
\refstepcounter{expno}{\small (\alph{expno})} & 128 & BIN & 2048 dims \iftoggle{tech_report}{&}{} (64$\times$) & 8 (2048 bits) & 7.63 MB & + 0.38 ms\\
\refstepcounter{expno}{\small (\alph{expno})} & 128 & BIN & 1024 dims \iftoggle{tech_report}{&}{} (128$\times$) & 4 (1024 bits) & \textbf{3.81 MB} & + 0.22 ms\\
\refstepcounter{expno}{\small (\alph{expno})} & 128 & PQ & 4 dims/sq \iftoggle{tech_report}{&}{} (16$\times$) & 32 & 30.5 MB & + 3.9 ms\\
\refstepcounter{expno}{\small (\alph{expno})} & 128 & PQ & 8 dims/sq \iftoggle{tech_report}{&}{} (32$\times$) & 16 & 15.3 MB & + 2.0 ms\\ \bottomrule
\end{tabular}
\end{adjustbox}
\end{center}
\iftoggle{tech_report}{%
\caption{\textbf{Dimensionality, storage requirements and computation time for all image representations.} The rows in this table correspond to those in Table~\ref{tab:res_tab}. Timings for compression methods are specified as additional time added to the total feature encoding time, and those in parenthesis indicate GPU timings where applicable.}
}{%
\caption{\textbf{Dimensionality, storage requirements and computation time.} The rows in this table correspond to those in Table~\ref{tab:res_tab}. Timings for compression methods are specified as additional time added to the total feature encoding time, and those in parenthesis indicate GPU timings where applicable.}
}
\label{tab:stat_tab}
\end{table}

\begin{figure}[t]
    \makebox[\textwidth][c]{\includegraphics[width=\widefigure\textwidth]{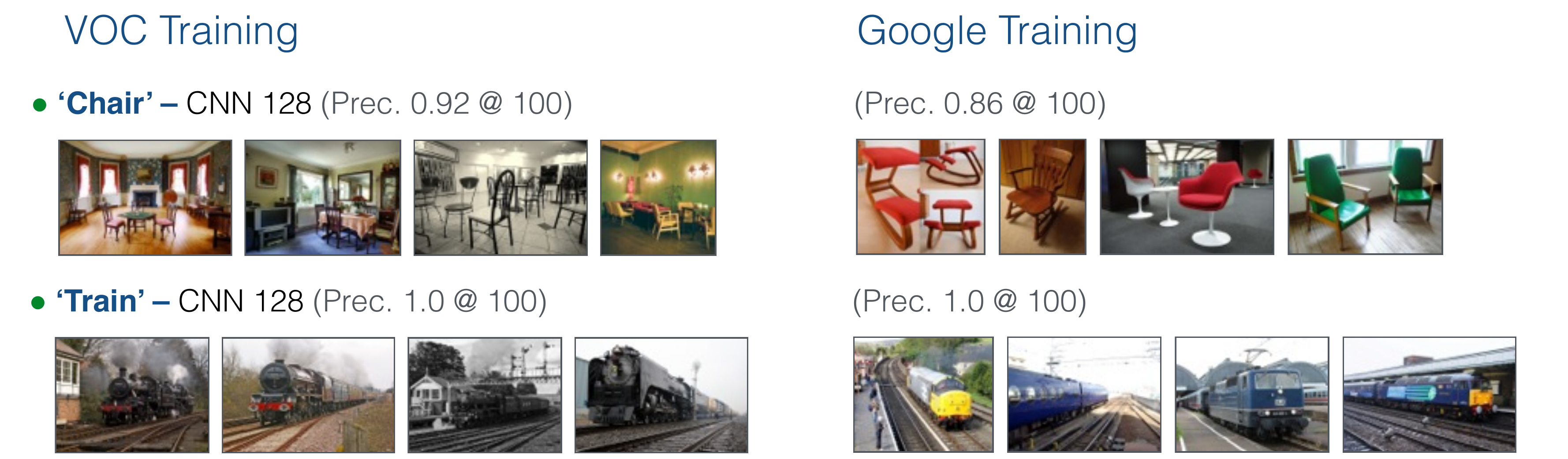}}
    \caption{\textbf{Difference between retrieved results when training using VOC data and Google training data.} Results are shown over the MIRFLICKR-1M dataset (Scenarios 2b and 3b).}
    \label{fig:google-train}
\end{figure}

The binary representations~(\ref{exp:cnn_128_rpbin2k} \& \ref{exp:cnn_128_rpbin1k}), combined with a linear SVM, also exhibit competitive performance despite a memory footprint lower even than the PQ-compressed codes.
The ranking of such features can also be significantly sped-up using hardware-accelerated Hamming distance computation. Nonetheless, the use of binary features requires a different ranking model, and so its application is left for future work.
%
%
The fact that the ConvNet features are very sparse,
with the CNN-128 representation typically being over $60\%$ zeros, is one reason
why they are so amenable to compression, and it is possible that with compression
methods geared specifically to capitalise on this sparsity even higher compression
ratios could be achieved.
\iftoggle{tight}{\vspace{-0.7em}}{}
\subsubsection{Scenario 2b (VOC Train/MIRFLICKR Test).}
Given that the MIRFLICKR-1M dataset contains many instances of all of the PASCAL VOC classes, moving
to testing solely on MIRFLICKR leads to a jump in performance of the results across all methods.
Nonetheless, this scenario provides a closer representation of the performance of a real-world
on-the-fly object category retrieval system, given that the image statistics of the MIRFLICKR-1M dataset
are not known in advance.
\iftoggle{tight}{\vspace{-0.7em}}{}
\subsubsection{Scenario 3a (Google Train/MIRFLICKR Test).}
Switching to noisy training images from Google rather than the pre-curated PASCAL VOC training images
as expected results in a small drop ($\sim 6\%$) across the board for all methods. However, the
precision at the top of the ranking remains subjectively very good. Nonetheless, as shown in
Figure~\ref{fig:google-train}, the actual images returned from the dataset are very different,
which reflects the differences in the training data sourced from Google Image search versus that
from the curated dataset. For example, a query for `chair' returns predominantly indoor scenes
with regular dining-table chairs when using VOC training data, and more avant-garde, modern
designs, generally centred in the frame when using Google training data.
\iftoggle{tight}{\vspace{-0.7em}}{}
\subsubsection{Scenario 3b (Google Train + negative pool/MIRFLICKR Test).}
In this scenario, we switch to using a fixed pool of negative data
sourced from a set of `negative' queries, and it can be seen how this
improves the results by up to $\sim 5\%$.
This may be a result of the larger negative training pool size
($\sim16,000$ images
\vs$\sim4,000$ images when using queries for all other VOC classes to provide the negative data as we do in Scenario 3a).
Given the assumed lack of coverage in
the fixed negative image pool (as it is sourced by issuing queries for
deliberately non-specific terms to facilitate its application to as broad a range of queries
as possible), this suggests that to a certain extent lack of diversity can be made up for by using
a larger number of negative training images.

\iftoggle{tight}{\vspace{-0.7em}}{}
\subsection{Implementation Details}
\label{sec:implement}
Our implementation of IFV and ConvNet image representations follows that of~\cite{Chatfield14}.
In more detail, for IFV computation we use their setting \mbox{`FK IN 512 (x,y)'}, which corresponds to:
(i)~dense rootSIFT~\cite{Arandjelovic12} local features with spatial extension~\cite{Sanchez12}, extracted with 3 pixel step over 7 scales ($\sqrt{2}$ scaling factor); 
(ii)~Improved Fisher vector encoding~\cite{Perronnin10a} using a GMM codebook with 512 Gaussians;
(iii)~intra normalisation~\cite{Arandjelovic13} of the Fisher vector.

Our ConvNet training and computation framework is based on the publicly available Caffe toolbox~\cite{Jia13}.
The two ConvNet configurations, considered in this paper (`CNN M 2048' and `CNN M 128') are pre-trained on the ImageNet ILSVRC-2012 dataset
using the configurations described in~\cite{Chatfield14}\footnote{\url{http://www.robots.ox.ac.uk/~vgg/software/deep_eval/}}.
Namely, they contain 5 convolutional and 2 fully-connected layers, interleaved with rectification
non-linearities and max-pooling. The stack of layers is followed by a 1000-way soft-max classifier, which is removed after pre-training is finished
(turning a ConvNet from an ImageNet classifier to a generic image descriptor). The only difference between the two ConvNets is the dimensionality of the second 
fully-connected layer, which is 2048 for `CNN M 2048' and 128 for \mbox{`CNN M 128'}.

In order to provide a similar setup to our on-the-fly architecture in Section~\ref{sec:arch},
which uses a linear predictor $\langle \bw,\enc(\image)\rangle$ learnt using SVM hinge loss
and a quadratic regulariser, as our learning stage we use a standard linear support vector
machine implementation. The $C$ parameter is determined using the VOC validation set for
scenario 1, and fixed at 0.25 for all other experiments.

\section{On-the-fly Architecture}
\label{sec:arch}

\iftoggle{tech_report}{%
\begin{figure}[t]
    \makebox[\textwidth][c]{\includegraphics[width=\narrowfigure\textwidth]{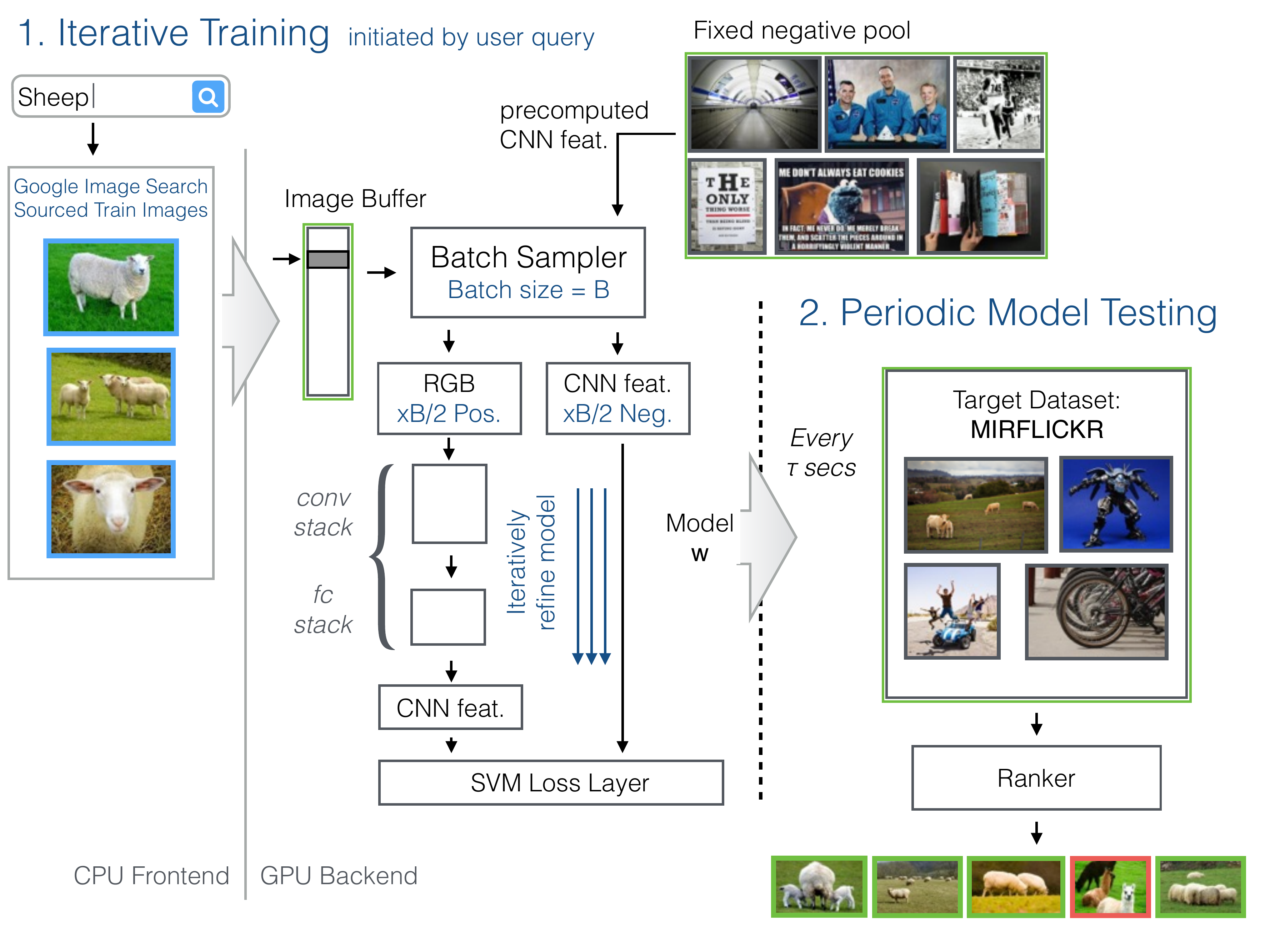}}
    \caption{\textbf{Architecture of our on-the-fly object category retrieval system.} The entire framework
    aside from the image downloader is resident on the GPU, with data stored in GPU memory outlined in
    green. Its operation is split into two stages: (i) iterative training, as initiated by a user text
    query and (ii) periodic model testing to obtain a ranking over the target dataset (refer to the
    text for further details).}
    \label{fig:architecture}
\end{figure}
}{}

Having evaluated various image representations in Sect.~\ref{sec:retrieval}, we now describe the architecture of the object category retrieval system, which fully exploits the advantages of ConvNet image
representations. From the user experience point of view, the main requirement to our system is instant response: the first ranking of the repository images should be obtained immediately (in under a second),
with a potential improvement over time. This dictates the following design choice: downloading the training images from the Internet should be carried out in parallel with training a model on the already
downloaded images in the on-line fashion. As a result, at any point of time, the current model can be used to perform ranking of the dataset images. 

For this approach to work, however, image representation should satisfy the following requirements: 
(i) highly discriminative, so that even a handful of training samples are sufficient to learn a linear ranking model;
(ii) fast-to-compute, to maximise the amount of training data processed within the allocated time budget;
(iii) low memory footprint, to allow for storing large-scale datasets in the main memory, and ranking them efficiently.
As has been demonstrated in Sect.~\ref{sec:retrieval}, a ConvNet image representation is a perfect match for these requirements. Indeed, pre-training on a large image collection (ImageNet) leads to highly
discriminative representation, and even a few training samples are sufficient for training an accurate linear model; ConvNet features can be computed very quickly on the highly-parallel GPU 
hardware; they have low dimensionality (even without PQ compression) and can be instantly scored using a linear model on the GPU.


Our on-the-fly architecture is illustrated in Fig.~\ref{fig:architecture}. It is divided into the CPU-based front-end (which controls the graphical user interface and downloads the training
images from the Internet) and the GPU-based back-end, which continually trains the ranking model on the downloaded images and periodically applies it to the repository. The category retrieval
is carried out as follows.

\noindent\textbf{Off-line (pre-processing).} To allow for fast processing, the ConvNet features for the target dataset images are pre-computed off-line, using the CNN-128 architecture.
We also prepare the fixed negative image pool for \emph{all queries} by issuing our negative pool queries (see Section~\ref{sec:scenarios}) to both Bing and Google image search, and downloading the returned
URLs.
The negative image feature features are also pre-computed.
The memory requirements for storing the pre-computed features are as follows: 488 MB for the MIRFLICKR-1M
dataset and 78 MB for the pool of 16K negative features. 
It is thus feasible to permanently store the features of both negative and dataset images in the high-speed
GPU memory even without compression of any kind (a consumer-grade NVIDIA GTX Titan GPU, used in our
experiments, is equipped with 6GB RAM). As noted in Section~\ref{sec:evdesc}, the ConvNet features can
be compressed further by up to $16\times$ using product quantization without significant degradation in
performance, making datasets of up to 160M images storable in GPU memory, setting 1GB aside for storage
of the model (compared to 10M images without compression), and more if multiple GPUs are used. Many
recent laptops are fitted with a GPU containing similar amounts of memory, making our system
theoretically runnable on a single laptop. Furthermore, whilst storing the target repository on the GPU
is preferable in terms of the ranking time, in the case of datasets of 1B+ images, it can be placed in the
CPU memory, which typically has larger capacity.

\noindent\textbf{On-line (CPU front-end).}
Given a textual query, provided by a user (\eg in a browser window), the front-end starts by downloading relevant images, which will be used as positive samples for the queried  category and fed to the GPU back-end. At regular time intervals, the front-end receives a ranked list of dataset images from the back-end, and displays them in the user interface.

\noindent\textbf{On-line (GPU back-end).}
The GPU back-end runs in parallel with the front-end, and is responsible for both training the ranking model and applying it to the dataset.
\emph{Training} an $L_2$-regularised linear SVM model is carried out using the mini-batch SGD with Pegasos updates~\cite{Shalev-Shwartz07}:
at iteration $t$, the learning rate is $\frac{1}{\lambda t}$, where $\lambda$ is the $L_2$-norm regularisation constant, set to $1$ in our experiments.
Each batch contains an equal amount of positive and negative samples; the total batch size was set to $B=32$ in our experiments.
The training commences as soon as the first positive image has been downloaded and
is received from the front-end, after which $B$ random crops are taken each
iteration from the pool of positive training images downloaded
so far. The front-end in the meantime will continue downloading
new images from the Internet, constantly increasing the size of the positive image
pool and the diversity of the extracted crops.
We note that while the positive image features need to be computed on-the-fly, this is very quick in the case of ConvNets.
\emph{Ranking} takes place using the current SVM model every $\tau$ seconds (we used $\tau\sim0.18$). As mentioned above, the pre-computed dataset features are pre-stored on a GPU,
so the scores for 1M images are computed in $\approx 0.01$s. The 1M scores are then ranked (also on GPU, $\approx 0.002$s) and the list of the top-ranked images is passed to the front-end
to be displayed to the user.
All components of the GPU back-end are implemented within the same framework, derived from Caffe~\cite{Jia13}.

\subsection{System Performance}\label{sec:arch-eval}


\begin{figure}[t]
    \makebox[\textwidth][c]{\iftoggle{tech_report}{\hspace{2.3cm}}{}\includegraphics[width=\pagefigure\textwidth]{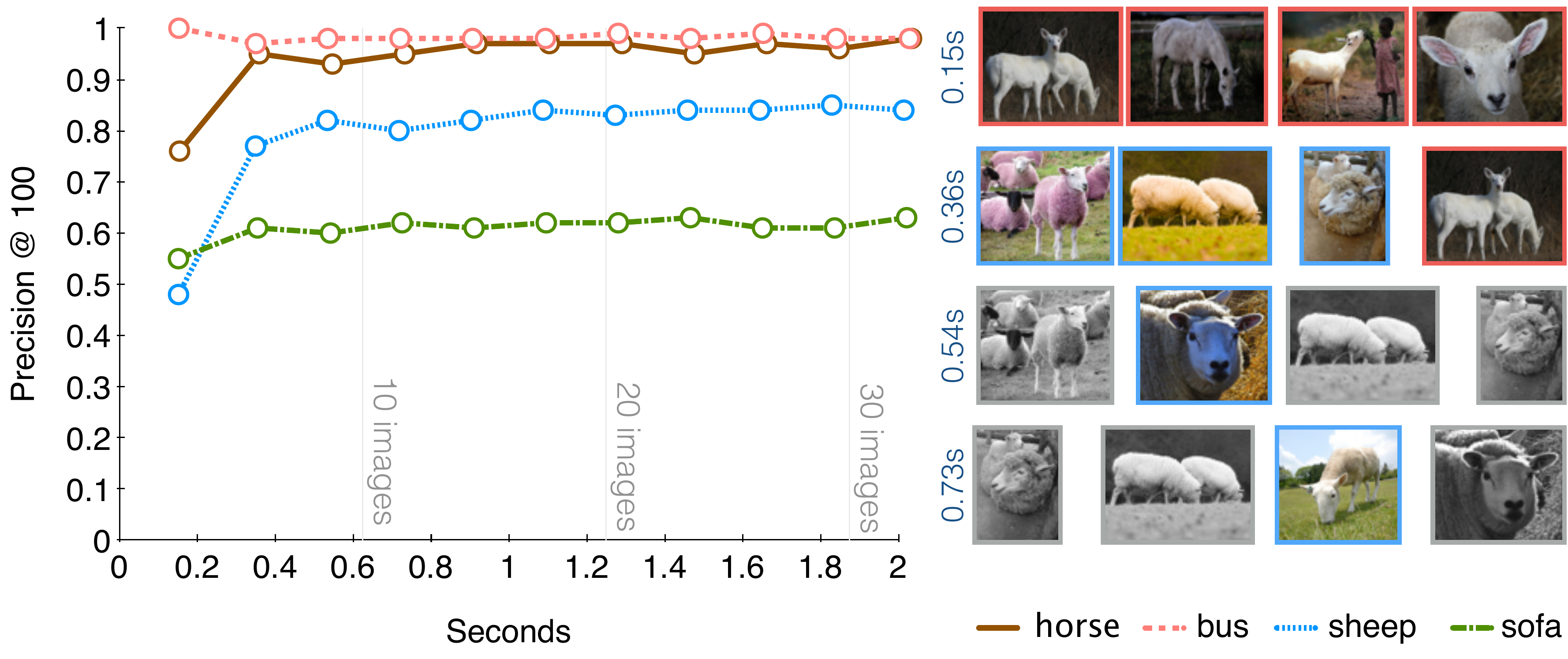}}
    \caption{\textbf{Precision @ 100 against training time for four queries using
    our on-the-fly architecture.}
    The number of images in the dynamically expanding positive image training pool
    over time is also marked on the plot. The top-4 returned images for the
    `sheep' query at the first four time-steps (up to 0.73s) is shown to the right.
    False positives are outlined in red, and new images in the top-4 at each time step
    are outlined in blue. Even for this moderately challenging query, the model
    settles in under a second.}
    \label{fig:otf-rankings}
\end{figure}

In order to evaluate the real-world performance of the system, we ran queries for
several PASCAL VOC classes and tracked how the performance (measured
in terms of Precision @ 100) evolved over time. To simulate the latency introduced
by downloading images from the Internet, we limited the rate of positive images
entering the network to 12 images/second (which is what we found to be a typical
average real-world rate on our test system). These images were sampled randomly
from the top-50 image URLs returned from Google Image search.

The results of these experiments for four classes are shown in Figure~\ref{fig:otf-rankings}. Even for some of the most challenging PASCAL VOC classes `sheep' and `softa', the performance converged to its final value in $\sim0.6$ seconds, and as can be seen from the evolving ranking at each time-step the ordering at the top of the ranking generally stabilizes within a second, showing a good diversity of results. For easier classes such as `aeroplane', convergence and stabilization occurs even faster.
%

In real terms, this results in a typical query time for our on-the-fly architecture,
from entering the text query to viewing the ranked retrieved images, of
\textbf{\mbox{1--2}~seconds~and~often~less} to complete convergence and stabilization
of results. However, one of the advantages of our proposed architecture is that it
is adaptable to differing query complexity, and we can return good results early
whilst still continuing to train in the background if necessary, exposing the
classifier to an expanding pool of training data as it is downloaded from the web and
updating the ranked list on-the-fly.



%


\iftoggle{tech_report}{%
\subsection{Impact of Training Image Count}\label{sec:impact}

\begin{figure}[p]
    \makebox[\textwidth][c]{\includegraphics[width=\narrowfigure\textwidth]{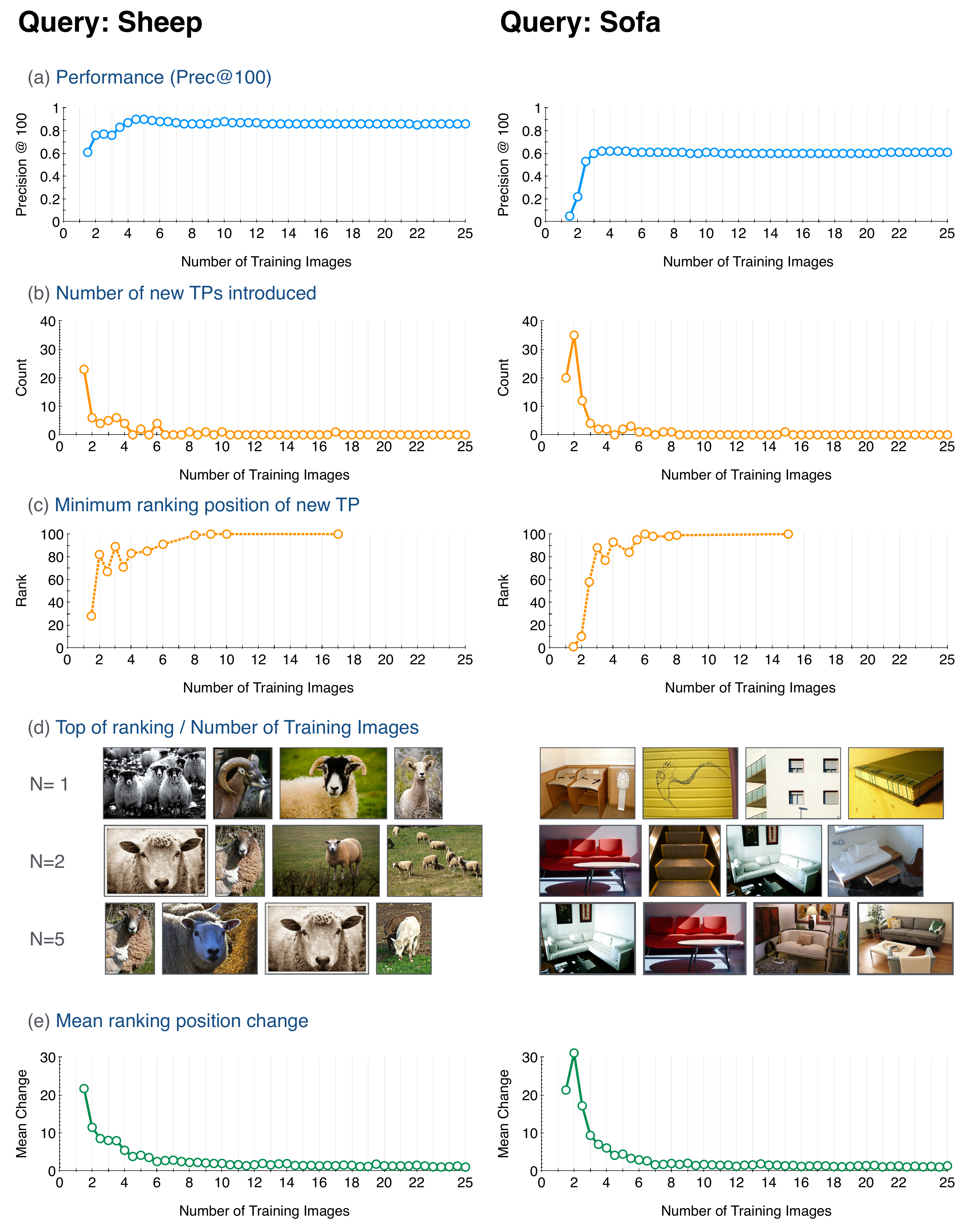}}
    \caption{\textbf{Evolution of performance with increasing number of positive training images}. Results
    are presented for two of the queries presented in Section~\ref{sec:arch} of the paper. (a) shows the performance (measured as Prec @ 100), (b) the number of new true positives entering
    the ranking (top 100) as the number of training images increases, (c) shows the minimum initial
    ranking position of any of those true positives, (d) shows the head of the ranked list after
    \emph{N}=1..5 images have been added, and (e) shows the mean change in ranking position for
    all images in the top 100.}
    \label{fig:rank-analysis}
    \iftoggle{tight}{\vspace{0.2em}}{}
\end{figure}

In this section, we present further more detailed analysis of the changes that
occur as more training images are fed into the network, to supplement those described
above for the two most challenging classes `sheep' and `sofa'.
The motivation is to determine the role of the size of the positive training image
pool in the performance of the system. Note that to this end the experimental setup is slightly different to the previous section, as after inputting each training image into the system we waited for the output classifier to stabilize.
We analyse the impact on each class in turn, referring to
Figure~\ref{fig:rank-analysis}.

Considering first the `sheep' class, with only a single training image 70\% of the
final performance (as measured by precision @ 100) is reached, and the top of the
ranked list contains many sheep. However, most of the highly ranked images are of
horned sheep, suggestive of the bias introduced by training only on a single image.
As the number of training images is increased to 2, the top-ranked images become
much more diverse, with this translating into a further final small jump in
performance as the third training image is fed into the network.

The `sofa' class provides an example of how the architecture deals with a more
challenging query, with a larger degree of intra-class appearance variation.
In this case, a single training image clearly does not suffice, as the ranked list
returned for a single training image has performance close to random, with no sofas
retrieved. However, this very quickly changes as a second image is fed into the
network, with 35 new true positives entering the top 100. Following this exposure,
the top retrieved images are greatly improved, mostly being of sofas. Feeding five
images into the network yields a further modest increase in diversity at the top
of the ranked list.

In both cases, for this dataset any new true positives introduced to the top 100
after the introduction of the third or fourth training images have a very high
initial position ($\sim80$) and the mean change in ranking position is very low
($\sim1.5$) suggesting that a coarse model can be trained with relatively few images,
and improvements after this time predominantly effect the tail of the ranked list.
This suggests that even when initially a very small number of training images are
available, a user interface where the head of the ranked list is presented to the user
almost immediately (trained on the small amount of training data which is available)
whilst training continues in the background to refine the tail of the ranked results
is possible. Such a restriction does not apply in our case, since as mentioned
in Section~\ref{sec:arch-eval} in general we have 30+ training images available to us
within a few seconds of launching a query.

}{}

\iftoggle{tech_report}{%
\subsection{Novel On-the-fly Queries}\label{sec:otf}
}{%
\iftoggle{tight}{\vspace{-1em}}{}
\subsubsection{Novel On-the-fly Queries.}\label{sec:otf}
}

\begin{figure}[t]
    \iftoggle{tech_report}{%
    \makebox[\textwidth][c]{\includegraphics[width=\widefigure\textwidth]{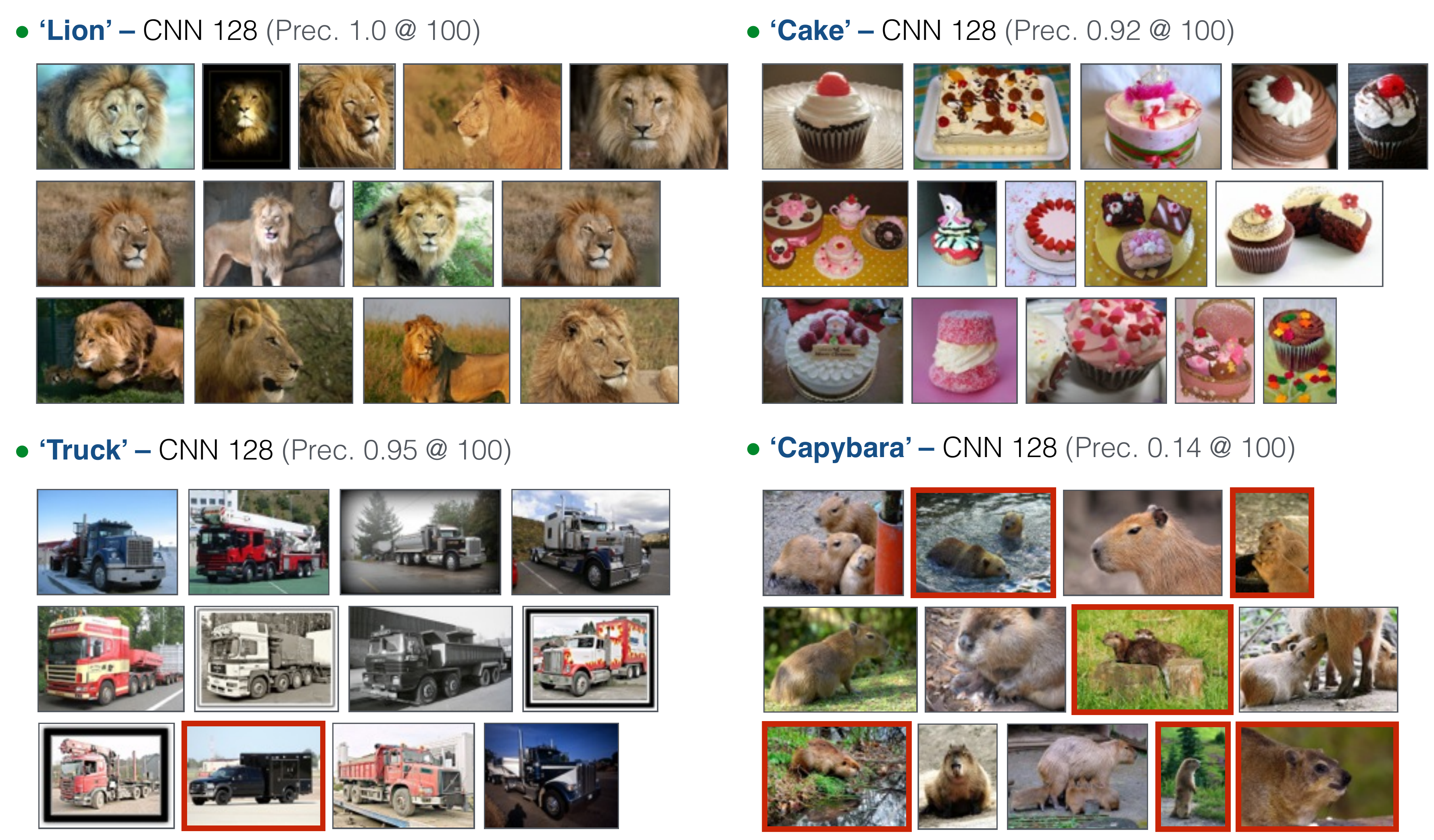}}
    }{%
    \makebox[\textwidth][c]{\includegraphics[width=\widefigure\textwidth]{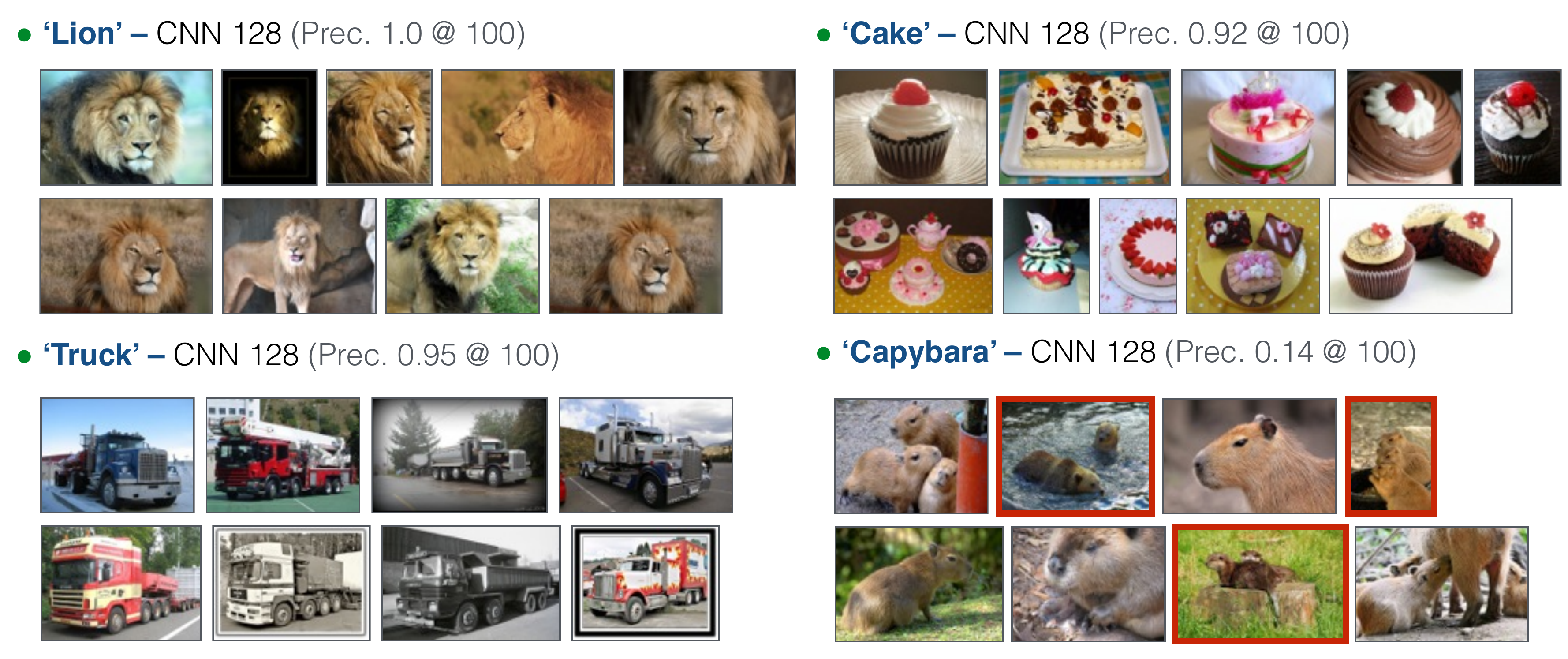}}
    }%
    \caption{\textbf{Sample results for queries outside of the twenty PASCAL VOC classes.} False positives are outlined in red.}
    \label{fig:extra-queries}
\end{figure}

Although experimental results have thusfar only been presented for the PASCAL VOC classes,
the advantage of an on-the-fly architecture is that no limitation is imposed on
the object categories which can be queried for, as a new classifier can be trained
on demand (in our case using Google Image search as a `live' source of training data).
We present some additional selected results of the on-the-fly system in
Figure~\ref{fig:extra-queries}, using the same setup as in \emph{Scenario 3b}
and query terms disjunct from the
\iftoggle{tech_report}{%
twenty PASCAL VOC classes to test its performance for such novel on-the-fly queries.
}{%
twenty PASCAL VOC classes.
}
It can be seen that the architecture is very much
generalisable to query terms outside of the PASCAL
category hierarchy.

\iftoggle{tech_report}{%
Some queries such as `lion' were particularly challenging for shallow feature
representations such as Fisher Kernel, due to the repeating thick fur pattern and
bushes present in many of the training images retrieving a large number of the
bursty images described in Section~\ref{sec:eval}. However, ConvNet-based
features appear to be much more robust to this effect, with precision~@~100 of 1.0.
The architecture is also capable of returning more abstract concepts such as
`cityscape' or `forest' in addition to more concrete objects such as `cake' and
`truck' (shown in the figure).

Finally, even when querying MIRFLICKR1M for the relatively obscure
`capybara'~(Figure~\ref{fig:extra-queries}), the returned false
positives all fit within a tight configuration of classes of very
similar appearance (`otter', `squirrel', `meercat') and, of course,
the composition
of the MIRFLICKR1M dataset is unknown, so it
could be that there are very few images of `capybara' in the
dataset.
}{}

\section{Conclusion}\label{sec:conclusion}

In this paper we have presented a system for on-the-fly object category retrieval, which builds upon the recent advances in deep convolutional
image representations. We demonstrated how such representations can be efficiently compressed
and used in a novel incremental learning architecture, capable of retrieval
across datasets of 1M+ images within seconds and running entirely on a
single GPU.

For larger datasets the CPU, or multiple GPU cards, could be employed for ranking
once the classifier has been learnt on the GPU. Along with further investigation of
how the diversity of the ranked results changes over time, this is the subject of
future work.


%
%
%


\iftoggle{tech_report}{%
\section*{Acknowledgements}
}{%
\iftoggle{tight}{\vspace{-0.7em}}{}
\paragraph{Acknowledgements}
}
This work was supported by the EPSRC and ERC grant VisRec no.\ 228180. We gratefully acknowledge the support
of NVIDIA Corporation with the donation of the GPUs used for this research.


\nottoggle{tech_report}{%
\bibliographystyle{splncs}
}{%
\bibliographystyle{IEEEtran}
}%
\bibliography{bib/longstrings,bib/shortstrings,bib/vgg_local,bib/vgg_other,bib/vgg_otf}

\end{document}